\documentclass{svproc}
\usepackage{url}
\usepackage{amsmath}
\usepackage{amssymb}
\usepackage{graphicx}
\usepackage{subcaption}
\usepackage{xcolor}
\usepackage{array}
\newcommand{\PreserveBackslash}[1]{\let\temp=\\#1\let\\=\temp}
\usepackage{multirow}
\newcolumntype{C}[1]{>{\PreserveBackslash\centering}p{#1}}
\newcolumntype{R}[1]{>{\PreserveBackslash\raggedleft}p{#1}}
\newcolumntype{L}[1]{>{\PreserveBackslash\raggedright}p{#1}}
\begin{document}
\mainmatter              
\title{MS-UMamba: An Improved Vision Mamba Unet for Fetal Abdominal Medical Image Segmentation}
\titlerunning{MS-UMamba}  
%
\author{Caixu Xu \inst{1} \and Junming Wei\inst{2,*}\and 
Huizhen Chen\inst{3,*} \and Pengchen Liang\inst{4} \and Bocheng Liang\inst{5} \and Ying Tan\inst{5} \and Xintong Wei\inst{2}}
\authorrunning{CaiXu Xu et al.} 
%
%
\institute{Guangxi Key Laboratory of Machine Vision and Intelligent Control, Wuzhou University, Wuzhou 543002, China
\and
College of Electronical and Information Engineering, Wuzhou University, Wuzhou 543002, China\\
\email{junmingwei0926@163.com}
\and
Department of Public Education, Wuzhou Medical College, Wuzhou 543199, China\\
\email{chenhuizhen139@163.com}
\and
School of Microelectronics, Shanghai University, Shanghai 201800, China
\and
Shenzhen Maternity and Child Healthcare Hospital, Southern Medical University, Shenzhen 518000, China
}
\maketitle              
\vspace{-10mm}
\begin{abstract}  
Recently, Mamba-based methods have become popular in medical image segmentation due to their lightweight design and long-range dependency modeling capabilities. However, current segmentation methods frequently encounter challenges in fetal ultrasound images, such as enclosed anatomical structures, blurred boundaries, and small anatomical structures. To address the need for balancing local feature extraction and global context modeling, we propose MS-UMamba, a novel hybrid convolutional-mamba model for fetal ultrasound image segmentation. Specifically, we design a visual state space block integrated with a CNN branch (SS-MCAT-SSM), which leverages Mamba's global modeling strengths and convolutional layers' local representation advantages to enhance feature learning. In addition, we also propose an efficient multi-scale feature fusion module that integrates spatial attention mechanisms, which Integrating feature information from different layers enhances the feature representation ability of the model. Finally,  we conduct  extensive experiments on a non-public dataset, experimental results demonstrate that MS-UMamba model has excellent performance in segmentation performance.
\end{abstract}
\section{Introduction}
Prenatal ultrasound is the most commonly used screening method for birth defects and can be used to diagnose whether a fetus has congenital diseases~\cite{1}. For example, the segmentation of the fetal abdominal key anatomical structures allows for the observation of the size, shape, and position of abdominal organs, thus enabling the assessment of fetal growth and development~\cite{2,3}. Currently, the diagnostic work of ultrasound sonographers often involves a certain degree of subjectivity, and this process heavily relies on the doctor's  own level of expertise~\cite{4}. In addition, the manual outlining of anatomical structure boundaries by the sonographer is also very time-consuming and may introduce errors, leading to missed or misdiagnosed cases. Therefore, exploring fully automated segmentation of key anatomical structures in the fetal abdominal plane is crucial.

With the development of artificial intelligence algorithms and computer vision, deep learning techniques have achieved significant success in the field of medical image  segmentation~\cite{5,6,7}. However, the segmentation of standard fetal abdominal planes faces greater difficulties and challenges. First, the fetal abdominal plane is accompanied by significant issues such as ambiguity, blurring, and deformation of organ boundary information, which can severely impact feature learning and model performance. Secondly, in the fetal abdominal plane, there is complete pixel overlap between some organs, which may lead to confusion in the model's judgment between the two targets, thereby severely affecting the segmentation accuracy and the overall performance of the model. Finally, the fetal abdominal plane contains highly similar anatomical structures, and the variations in the size, position, and shape of these structures across different gestational weeks are substantial. These factors significantly increase the difficulty of feature learning and the segmentation of these anatomical structures.

To address the aforementioned challenges, we propose a novel  end-to-end automatic semantic segmentation model based on a hybrid convolutional-mamba framework (MS-UMamba),  aimed at enhancing the model's understanding and learning of anatomical structure features. Specifically, we propose a novel improved Visual State Space Block (SS-MCAT-SSM) that  introduces a CNN branch, enabling it to not only effectively capture contextual information but also efficiently extract local feature details. In addition, we also propose an efficient multi-scale feature fusion module that integrates spatial attention mechanisms, which integrates feature information from different layers. By leveraging contextual features and local features to enrich the semantic information in the feature maps,  enhances the feature representation ability of the model. Our contributions are as follows:
\begin{itemize}
    \item We propose a fetal ultrasound image semantic segmentation model  that combines CNN and Mamba Block. This model is capable of accurately segmenting six key organs in the standard fetal abdominal plane.
    \item We designed the SS-MCAT-SSM block and the ADFF multi-scale feature fusion block to further enhance the feature extraction efficiency of the MS-UMamba model. Experimental results show that these blocks significantly improve the segmentation performance of abdominal anatomical structures.
    \item We conduct extensive experiments on a non-public dataset. Experimental results demonstrate that the MS-UMamba model exhibits outstanding performance in segmentation, particularly in the segmentation of small targets.
\end{itemize}
\section{Related Work}
\subsection{Medical image segmentation}
Deep learning-based methods are becoming increasingly popular in medical image segmentation. For example, U-Net and its variant models have always been the mainstream in medical image segmentation~\cite{10,11,12}. Subsequently, several novel improved networks based on attention mechanisms, Transformer, and CNN + Transformer hybrid frameworks have been developed~\cite{13,14,15,16,17}. The emergence of the Mamba framework~\cite{18} has provided a new idea for research in medical image segmentation models. Ma et al.~\cite{19} proposed the U-Mamba model for 2D/3D biomedical image segmentation by designing a hybrid block combining CNN block and SSM block. Xing et al. ~\cite{20} proposed the SegMamba model, exploring the application of the visual Mamba network in 3D medical images. Ruan et al.~\cite{21} proposed VM-UNet, which replaces all convolutional blocks in the original U-Net model with Mamba blocks, achieving promising results on open medical image datasets. Subsequent works have further explored the application of the Mamba architecture in medical images~\cite{22,23,24,25}. However, this research primarily focus on the application of long-range dependency modeling, overlooking the relationship between contextual information and local features, as well as the fusion of multi-scale semantic information. 

\subsection{Fetal ultrasound image segmentation}
Significant progress has also been made in the research of semantic segmentation models for fetal ultrasound images~\cite{26,27,28,29}. Wu et al.~\cite{30} proposed a fully automated quality control and biometric measurement model for the fetal abdominal plane, which assists obstetricians in identifying the spine (SP), stomach bubble (SB), umbilical vein (UV), and measuring abdominal circumference biological parameters. Oghli et al. ~\cite{31} introduced attention gates and a multi-feature pyramid into the U-Net model, proposing the MFP-Unet network model for the automatic segmentation of anatomical structures such as biparietal diameter (BPD), head circumference (HC), abdominal circumference (AC), and femur length (FL) in ultrasound images. Liu et al.~\cite{32} introduced methods such as attention fusion and filters, proposing the AFG-net network model. Alzubaidi et al.~\cite{33} proposed an end-to-end automated segmentation model, fetSAM, which can segment the background, fetal brain, cavum septi pellucidi (CSP), and lateral ventricles (LV) in fetal head ultrasound images. However, this research focuses on  single-target segmentation,  without the need to consider the impact of issues such as pixel overlap, blurred boundaries, and other challenges between multiple segmentation targets on model performance.

\section{Method}
\subsection{Architecture Overview}
\begin{figure}[htbp]
\centerline{\includegraphics[width=\textwidth]{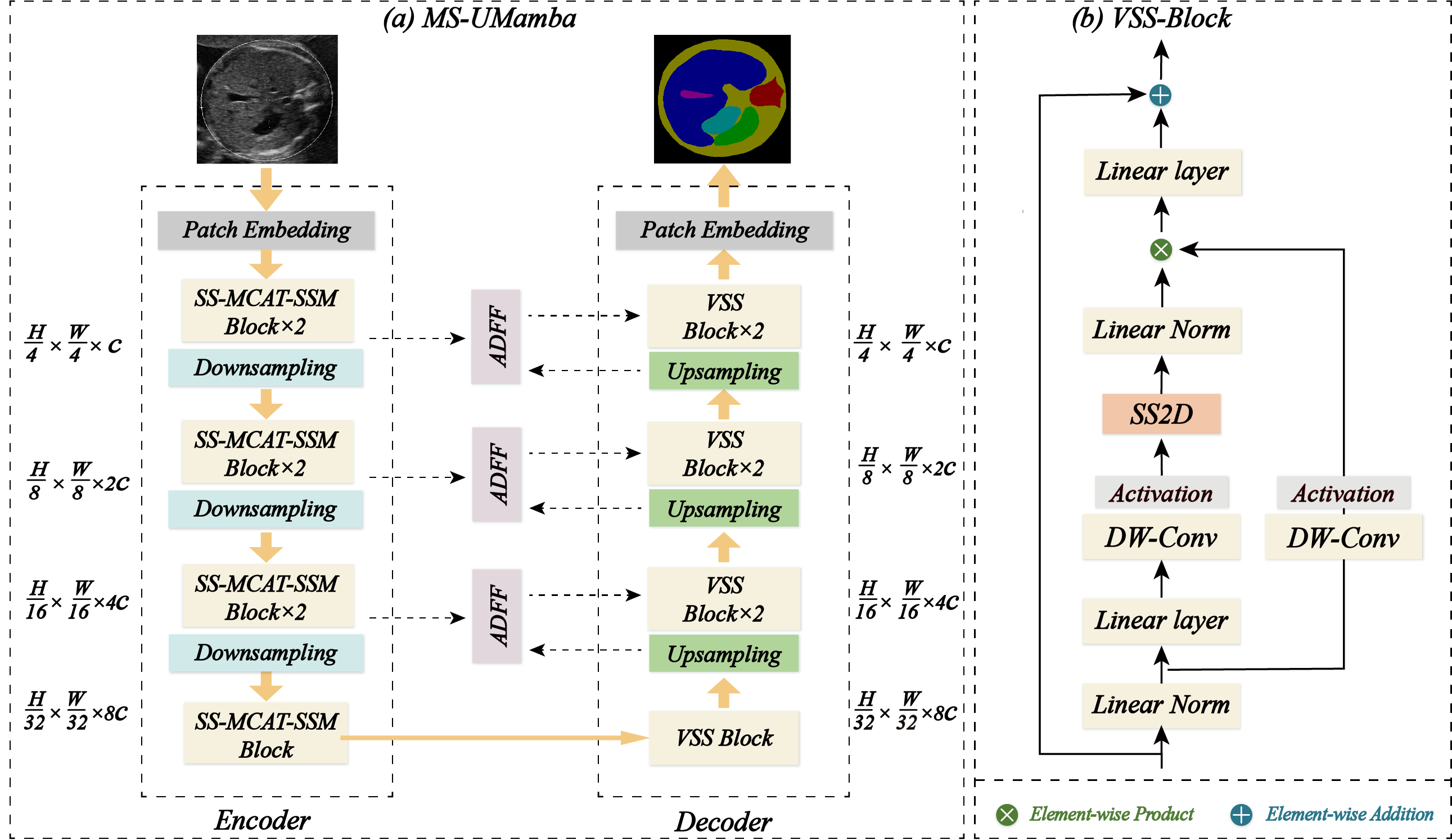}}
\caption{The illustration of MS-UMamba architecture.}
\label{fig1}
\end{figure}

The Overview of the proposed MS-UMamba model is shown in Fig.~\ref{fig1}. Our MS-UMamba composed of Patch Embedding, Mamba encoder, patch merging, feature fusion modules, Mamba decoder, patch expanding, and skip connections. Firstly, we use the Patch Embedding layer to map the given input ultrasound image $X\in R^{H\times W\times 3}$ to a feature map with dimensions H/4, W/4, and 96 channels. Next, the encoder extracts features of the specified dimensions through four stages. Specifically, each stage consists of three concatenated SS-MCAT-SSM blocks, after feature extraction using this improved Mamba structure, the output features are mapped to a deeper feature space, where the width and height are halved, and the number of channels is doubled, by the patch merging layer. Similarly, the decoder also consists of four stages. Each stage includes two VSS modules and a Patch Expanding module for upsampling, the width and height of the feature maps are doubled at each layer, while the number of channels is halved. In addition, we designed a multi-scale feature fusion module between the encoder and decoder, which uses skip connections to deeply fuse features from both the encoder and decoder. Finally, we use a 1×1 convolution to project the class of each pixel in the feature map after the decoder, enabling precise segmentation predictions.
\vspace{-6mm}
\subsection{SS-MCAT-SSM Blcok}
\vspace{+2mm}
\begin{figure}[htbp]
\centerline{\includegraphics[width=\textwidth]{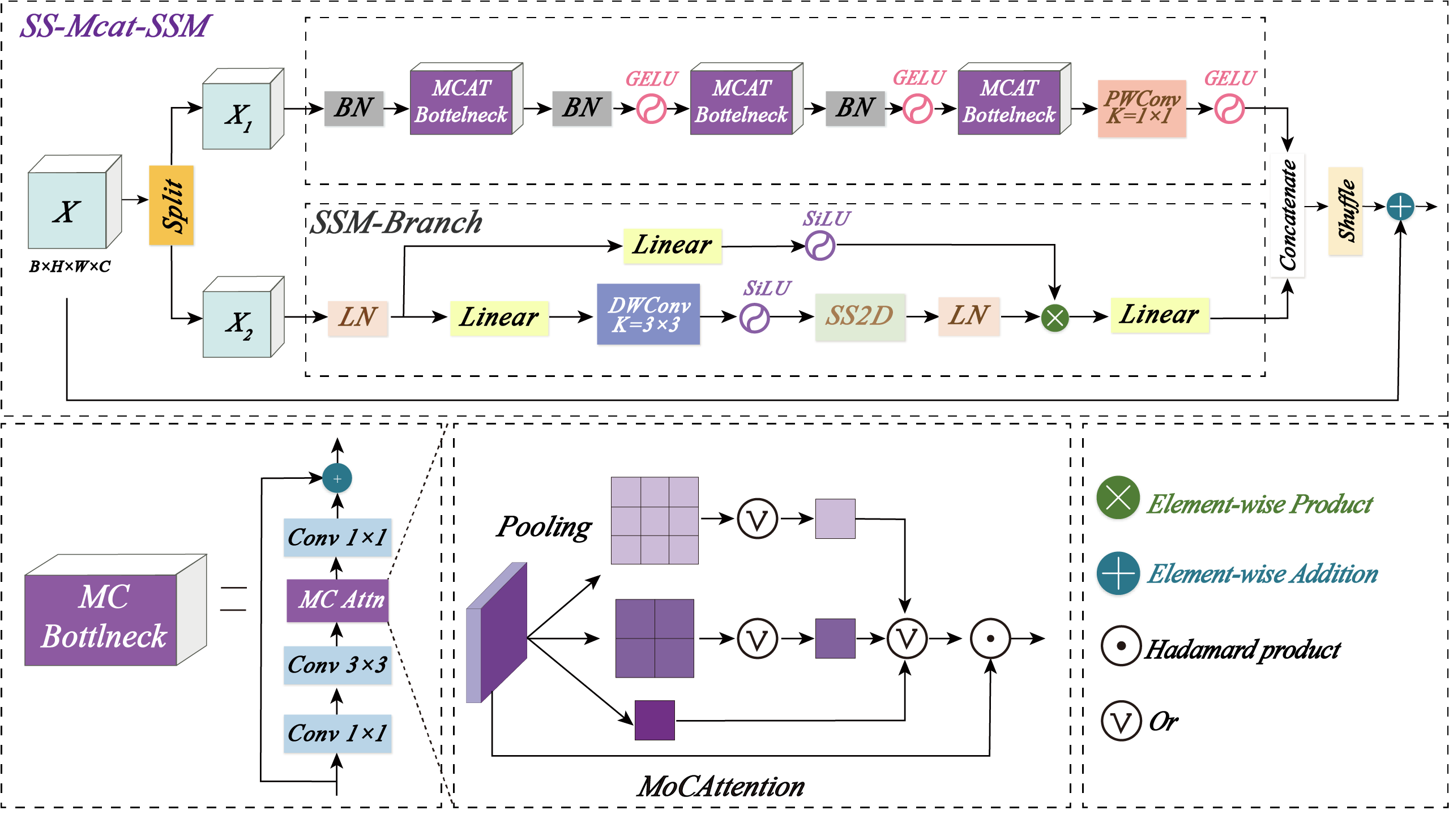}}
\caption{Design Details of the SS-MCAT-SSM Module.}
\label{fig2}
\end{figure}
The model based on the Mamba framework can effectively capture long-range dependencies, but it often lacks the ability to perceive local feature information. Inspired by the MedMamba model ~\cite{34}, we propose the dual-branch SS-MCAT-SSM block the backbone network feature extraction module (show in Fig.~\ref{fig2}). First, the input features are divided into two groups with equal channels using the channel-split method. Then, We input these two feature groups into the CNN-branch and Mamba-branch, respectively, for feature extraction. Finally, the output features from both branches are merged into a feature map with the same scale as the original input features using the channel-concat and channel-shuffle methods. Compared to the MEDMamba block, we have introduced a custom-designed BottleNeck block with an integrated Monte Carlo attention mechanism in the Conv-Branch. This enables the model to better extract local feature information, enhancing its adaptability to small target segmentation in medical images. The mathematical description of the improvements made to the McatBottleneck is as follows:
\begin{align}
x^{\prime} & = \text{Conv}_{1 \times 1}(x),  x\in \mathbb{R}^{B \times H \times W \times \frac{2}{C}}, \\
x^{\prime \prime} &= A_m(x^{\prime}) = \sum_{i=1}^{n} P_1(x, i) f(x, i), \\
x_{\text{out}} &= \text{Conv}_{1 \times 1}\left(\text{Conv}_{3 \times 3}(x^{\prime \prime})\right) + x.
\end{align}
Where $A_m(x)$ is the output attention map of Monte Carlo attention, where i represents the size of the output feature map after attention processing. $f(x,i)$ denotes the average pooling function, and $P_1(x,i)$ represents the association probability.
\subsection{Attention-Based Dynamic Feature Fusion Blcok}
\begin{figure}[htbp]
\centerline{\includegraphics[width=\textwidth]{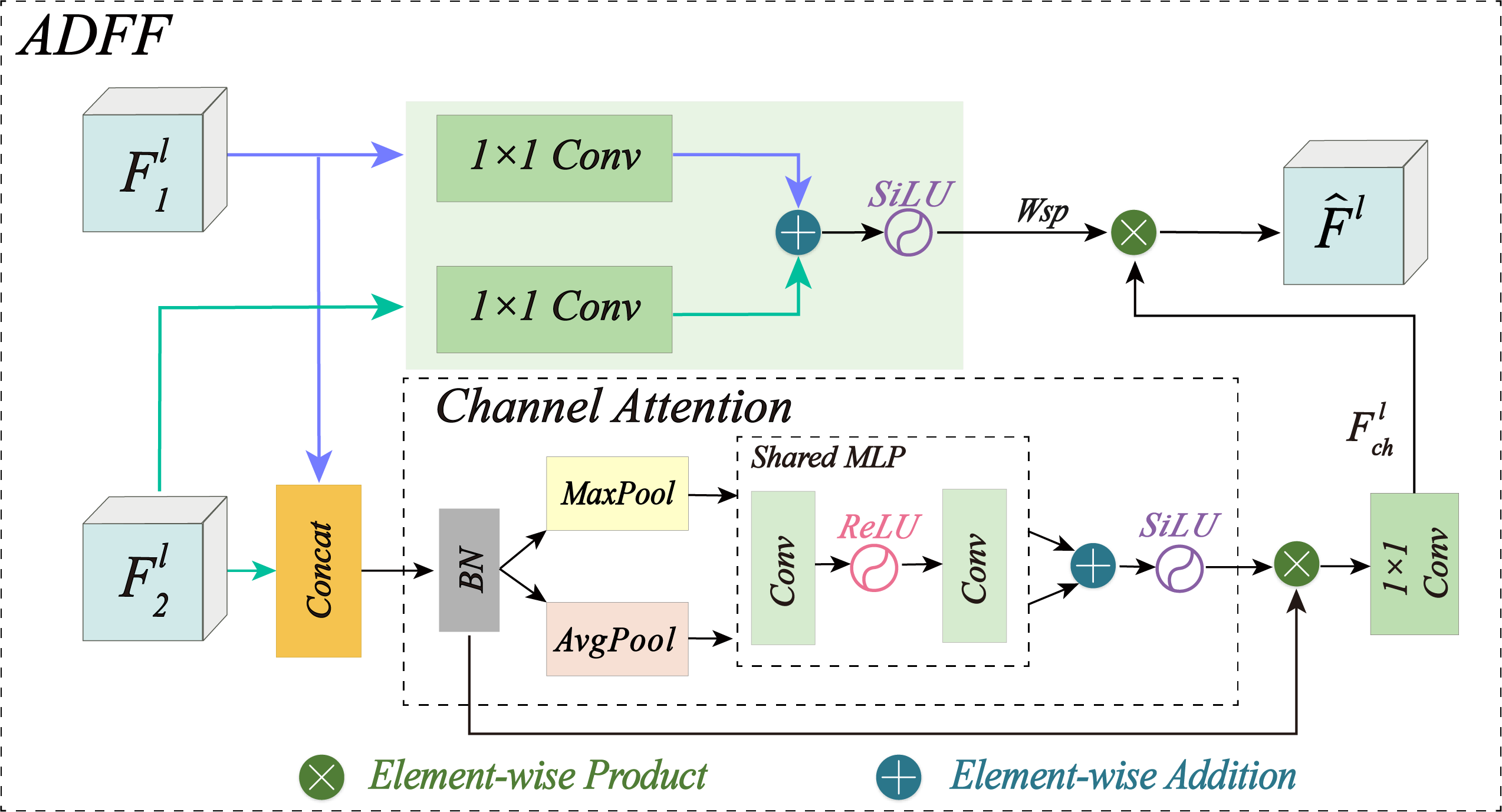}}
\caption{Design Details of the Attention-Based Dynamic Feature Fusion Module.}
\label{fig3}
\end{figure}
Reference~\cite{35} demonstrated that the DFF module an adaptively fuse multi-scale local feature maps based on global information. As shown in Fig.~\ref{fig3}, on this basis, we propose the ADFF module, which specifically incorporates an optimized channel attention mechanism. This allows the model to dynamically adjust the importance of each channel, thereby enabling more effective fusion of feature information from both the encoding and decoding stages. The features
$F_{1}^{l}  \in \mathbb{R}^{H\times W\times C}$
from the encoder and the feature map 
$ F_{2}^{l}  \in \mathbb{R}^{H\times W\times C} $ 
from the decoder are mapped to the specified feature dimensions through a 1×1 convolution layer, and then combined by summation. Then, the spatial dependencies of the local feature maps are modeled, and the global information is captured by activating through the Sigmoid function.
\begin{eqnarray}
\omega _{sp}=Sgmiod(Conv_{1}(F_{1}^{l})\oplus  Conv_{1}(F_{2}^{l})),
\end{eqnarray}
In addition, the feature maps $F_{1}^{l}$ and $F_{2}^{l}$ are concatenated to obtain a fused feature map, which is then processed through batch normalization to improve training reliability and stability. Then, the input feature map is processed using both global max pooling and global average pooling, resulting in two feature maps with different dimensions. The feature maps share a single multilayer perceptron (MLP) network, where each map undergoes feature dimensionality reduction and expansion through separate fully connected layers. Then, the two feature maps are stacked along the channel dimension, passed through a Sigmoid activation function, and multiplied element-wise with the input feature map to obtain the output feature map. Finally, we use a 1×1 convolution to map the features to the specified channels, enabling information interaction between this feature extraction module and others. The mathematical expression for this process is as follows:
\begin{align}
F^{l} &= \text{BatchNorm}\left(\text{ConCat}(F_{1}^{l}, F_{2}^{l})\right), \\
\omega_{ch} &= \text{Sigmoid}\left(\text{Conv}\left(\text{Pool}_{\text{Avg}}(F^{l}) + \text{Pool}_{\text{Max}}(F^{l})\right)\right), \\
F_{ch}^{l} &= \text{Conv}\left(\omega_{ch} \otimes F^{l}\right).
\end{align}

Ultimately, the information from both feature fusion branches is combined to enhance the interpretability of complex features.
\begin{eqnarray}
\hat{F} ^{l} =\omega _{sp}\otimes F_{ch}^{l} ,
\end{eqnarray}

Where $F^l$ is the output feature map after concatenation and batch normalization. $\omega _{ch}$ represents the channel weights at the l-th layer.$\hat{F} ^l$ is the final weighted feature map, which integrates the feature information from the two branches.

\subsection{Loss fuction}
In response to issues such as sample imbalance, overlapping anatomical structures, and small anatomical structures in ultrasound image segmentation tasks, our model constructs the loss function by combining the Focal loss function and Dice-Loss function, enabling it to more accurately reflect the differences between the segmented prediction structure and the ground truth labels. The Focal loss function dynamically adjusts the loss weights for different samples, ensuring that samples that are harder to distinguish receive greater attention. Its mathematical expression is as follows:
\begin{eqnarray}L_{Focal} & = & -\alpha (1-\hat{P} )^{\gamma }y log(\hat{P})-(1-\alpha )\hat{p}^{y}(1-y)log(1-\hat{p} ),
\end{eqnarray}
the Dice Loss function assigns higher weight to the loss of small targets during optimization, allowing the model to more easily capture the feature information of small structures during the learning process. Its mathematical expression is as follows:
\begin{eqnarray}
L_{Dice}=1-\frac{2\times|Y\cap \bar{Y} | }{|Y|+|\bar{Y}|} , 
\end{eqnarray}
where $Y$ represents the ground truth labels, $\bar{Y}$ denotes the predicted probabilities, $Y\cap \bar{Y}$ is the size of the intersection between the ground truth labels and the predicted results, $|Y|$ indicates the size of the ground truth labels, and $|\bar{Y}|$ represents the size of the predicted results.
\section{Experiments and Results}
\subsection{Experimental Setting}
The fetal abdominal ultrasound images utilized in this study were acquired from Shenzhen Maternal and Child Health Hospital. A total of 696 fetal abdominal ultrasound images, encompassing gestational ages ranging from 14 to 28 weeks. Six critical anatomical structures within the images were manually annotated by experienced ultrasound technicians. The distribution of these anatomical structures is detailed in Table~\ref{table1}. All experiments were conducted on a Linux server running Ubuntu 22.04 LTS, equipped with an Intel(R) Xeon(R) Platinum 8352V CPU operating at 2.10GHz and four NVIDIA GeForce RTX 3090 GPUs. Additionally, during the training process, we applied data augmentation techniques such as horizontal flipping with a 50\% probability, random scaling, padding with gray bars, and color space transformations. These augmentations increased the diversity and complexity of the dataset, thereby enhancing the robustness of the model.
\begin{table}[!ht]
    \centering
    \setlength{\tabcolsep}{12pt} 
    \caption{The distribution of anatomical structures within the dataset.}
    \begin{tabular}{c c c c}
    \hline
        Item    & Train  & Validation  & Test  \\ \hline
        SP      & 621    & 177  & 90  \\ 
        SL      & 503    & 144  & 72  \\ 
        AW      & 552    & 158  & 78  \\ 
        LV      & 547    & 156  & 79  \\ 
        ST      & 487    & 139  & 70  \\ 
        UV\&PV  & 548    & 157  & 79  \\ \hline 
    \end{tabular}
    \label{table1}
\end{table}
\vspace{-6mm}  
\subsection{Evaluation Metrics}
To provide a more comprehensive evaluation of the model's performance, we utilized five distinct evaluation metrics: Dice Similarity Coefficient (DSC), Intersection over Union (IoU), Precision (Pre), Sensitivity (Sen), and Specificity (Spe). Higher values for each metric indicate superior segmentation performance by the model. The definitions of these metrics are provided below:
\begin{align}
\mathit{Dice} = \frac{2 \times TP}{2 \times TP + FP + FN},~IoU = \frac{Area_{Pc} \cap Area_{Gc}}{Area_{Pc} \cup Area_{Gc}},
\end{align}
\begin{align}
Pre = \frac{TP}{TP + FP}, \quad Sen = \frac{TP}{TP + FN}, \quad Spe = \frac{TN}{TN + FP}.
\end{align}
Where TP denotes the number of correctly segmented key anatomical structures, FP indicates the number of falsely segmented key anatomical structures, and FN represents the number of key anatomical structures that were not segmented. 
\subsection{Comparison Experiment}
This paper presents the experimental comparison results of the MS-UMamba model with other state-of-the-art semantic segmentation models on the fetal abdominal ultrasound slice dataset, as shown in Table~\ref{table2}. The experimental results demonstrate that the MS-UMamba model achieves excellent performance in terms of mIoU, mDC, mSE, mSP, and mPRE with values of 67.62\%, 79.82\%, 84.78\%, 99.30\%, and 76.09\%, respectively, outperforming all other models in these metrics. Although the mPRE value is slightly lower than that of other models, the MS-UMamba model outperforms the comparison models significantly in all other metrics. Specifically, compared to the second-best performing VM-UNet model, the MS-UMamba model improves the mIoU, mDice, and mSE metrics by 1.97\%, 1.54\%, and 1.19\%, respectively.
\begin{table}[!ht]
    \centering
    \setlength{\tabcolsep}{8pt} 
    \caption{Quantitative comparison with previous SOTA methods.}
    \begin{tabular}{c c c c c c c}
    \hline
        Method           & Year & mIoU$\uparrow$ &mDC$\uparrow$ &mSE$\uparrow$  & mSP$\uparrow$    & mPRE$\uparrow$  \\ \hline
        VM-UNet          & 2024 & \underline{65.65}   & \underline{78.28}  & \underline{83.59}  & 99.16  & 74.41 \\ 
        SegMenter        & 2021 & 64.36   & 76.90  & 77.64  & \underline{99.28}  
        & \textbf{76.54} \\ 
        Segformer        & 2021 & 62.56   & 75.69  & 75.54  & 98.96  & 76.05 \\ 
        Swim-TransFormer & 2021 & 65.29   & 77.75  & 80.69  & 99.13  & 75.47 \\ 
        DeepLabv3+       & 2018 & 64.25   & 76.96  & 78.05  & 99.23  & 76.10 \\ 
        UperNet          & 2018 & 61.15   & 74.13  & 72.59  & 99.13  & \underline{76.17} \\ 
        MS-UMamba (Ours) & 2025 & \textbf{67.62}   & \textbf{79.82}  & \textbf{84.78}  
        & \textbf{99.31}  & 76.11 \\ \hline
    \end{tabular}
    \label{table2}
\end{table}

The performance of the MS-UMamba model in segmenting different anatomical structures is shown in Table~\ref{table3}. The model performs well in segmenting larger targets, with the highest IoU and Dice coefficients for LV, demonstrating strong segmentation capability. In contrast, when segmenting smaller targets such as UV, the model excels in specificity, effectively reducing false positives, thereby improving segmentation accuracy and the reliability of the results. In the future, the focus will be on optimizing the segmentation of SL and UV to enhance overall performance.
\vspace{-2mm}
\begin{table}[!ht]
    \centering
    \setlength{\tabcolsep}{8pt} 
    \caption{The performance of segmentation for different key anatomical structures.}
    \begin{tabular}{c | c c c c c}
    \hline
        Item & IoU$\uparrow$ & DC$\uparrow$ & SE$\uparrow$ & SP$\uparrow$ & PRE$\uparrow$  \\ \hline
        SP     & 62.06   & 76.59  & 80.30  & \underline{99.79}  & 73.21 \\ 
        SL     & 51.89   & 68.33  & 78.85  & 99.72  & 60.28 \\ 
        AW     & 63.51   & 77.69  & 73.21  & 98.84  & \underline{82.74} \\ 
        LV     & \textbf{77.17}   & \textbf{87.12}  & \textbf{90.77} & 98.92  & \textbf{83.74} \\ 
        UV     & 55.03   & 70.99  & 84.21  & \textbf{99.83}  & 61.36 \\ 
        ST     & \underline{64.50}   & \underline{78.42}  & \underline{86.64}  & 99.75  & 71.62 \\ \hline 
    \end{tabular}
    \label{table3}
\end{table}
\vspace{-6mm}
\subsection{Ablation Experiment}
We conducted a series of ablation experiments to validate the impact of the proposed SS-MCAT-SSM Block and ADFF Block on the model's segmentation performance. Using VM-Unet as the baseline model, we evaluated the contribution of each module to the task by progressively adding the corresponding modules. The specific results are shown in Table~\ref{table4}. 
\begin{table}[!ht]
    \centering
    \setlength{\tabcolsep}{6pt} 
    \caption{Ablation Results of Model Optimization.}
    \begin{tabular}{c c c c | c c c c c}
    \hline
        MED & MCAT & DFF & ADFF & mIoU$\uparrow$ 
        & mDC$\uparrow$  & mSE$\uparrow$  & mSP$\uparrow$  & mPRE$\uparrow$  \\ \hline
        $\checkmark$     &$\times$    & $\times$  & $\times$  & 66.54  & 78.94 & \underline{84.20}    & 99.26 & 75.11\\ 
        $\times$     & $\checkmark$   & $\times$  & $\times$  & \underline{66.84}  & \underline{79.20} & 83.88    & \underline{99.28} & 75.70\\ 
        $\times$     & $\times$   & $\checkmark$  & $\times$  & 53.54  & 66.98 & 70.22    & 98.53 & 64.80\\ 
        $\times$     & $\times$   & $\times$  & $\checkmark$  & 66.72  & 79.10 & 83.54           & \underline{99.28} & \underline{75.73}\\ 
        $\times$     & $\checkmark$   & $\times$  & $\checkmark$  & \textbf{67.62} 
        & \textbf{79.82}     & \textbf{84.78}   & \textbf{99.31}  & \textbf{76.11}\\ \hline
    \end{tabular}
    \label{table4}
\end{table}
When replacing the MedMamba backbone network with the Baseline backbone network, the model's segmentation performance was average, with an mIoU of 66.54\% and an mDice of 78.94\%. After incorporating the SS-MCAT-SSM module we designed, the model's segmentation performance significantly improved, with the mIoU increasing to 66.84\% and the mDice rising to 79.20\%. This result indicates that the SS-MCAT-SSM module enhances the model's ability to capture the characteristics of medical images, thereby effectively improving segmentation performance. Moreover, when the ADFF multi-scale fusion module was further incorporated, the segmentation model achieved optimal segmentation performance across all anatomical structures, with the mIoU metric increasing to 67.62\% and the mDice rising to 79.82\%, both representing the highest values. However, when the basic DFF module was introduced, the model's segmentation performance noticeably declined, with all metrics being lower than those in the other ablation experiments. The above results suggest that the combination of MCAT and ADFF effectively enhances feature extraction and fusion capabilities, enabling the model to achieve better segmentation performance on more complex anatomical structures.
\begin{figure}[htb]
\centerline{\includegraphics[width=\textwidth]{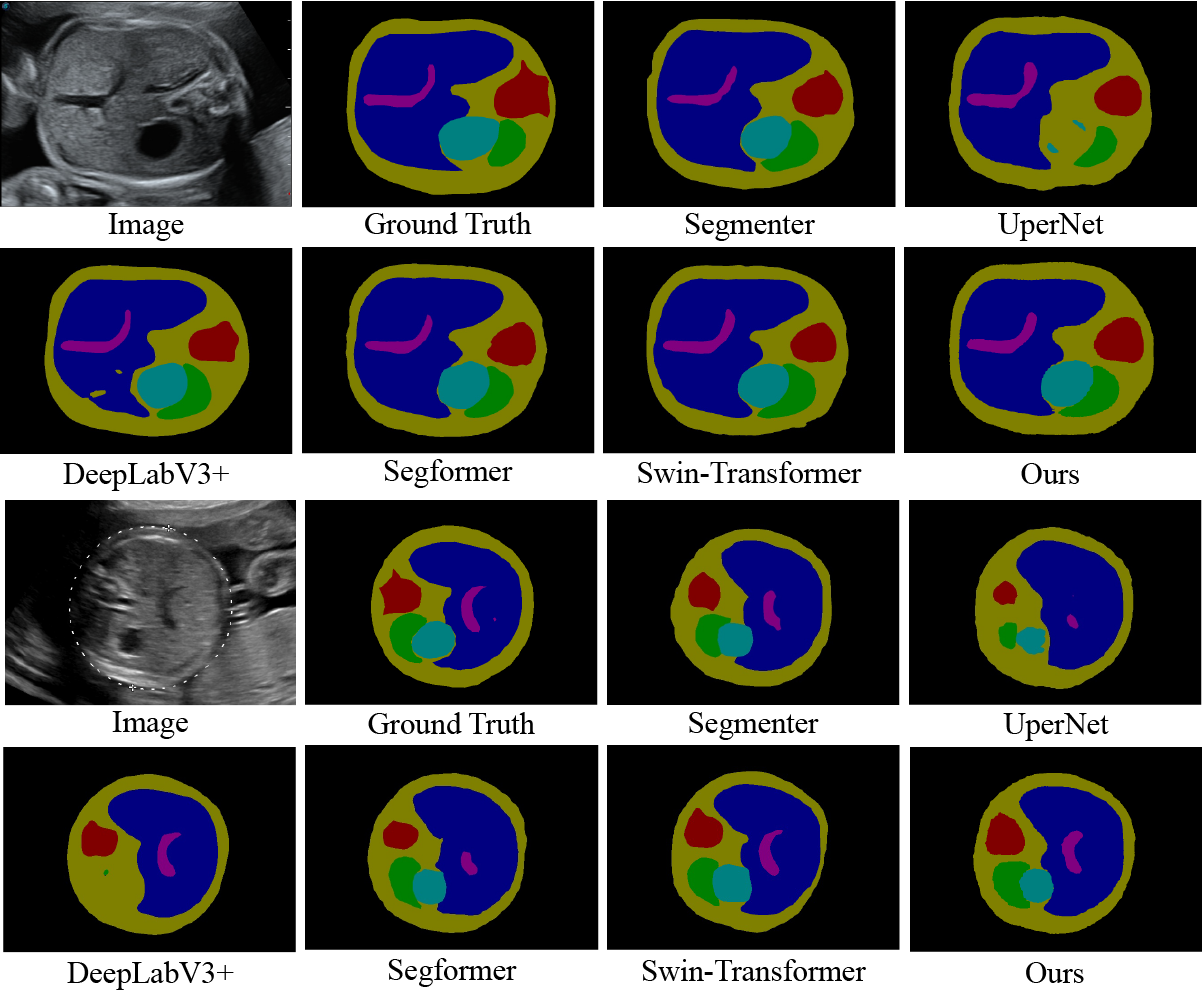}}
\caption{Visualization of segmentation results. The red, yellow, yellow-green, chartreuse, cyan, and azure masks indicate the SP, SL, AW, LV, UV, and ST respectively.}
\label{fig4}
\end{figure}
\subsection{Visualization Analysis}
Figure~\ref{fig4} presents the visualized segmentation results of the MS-UMamba model and several baseline models on the fetal ultrasound abdominal cross-sectional images. We observed that the MS-UMamba model performs excellently in terms of segmentation accuracy and boundary precision across all anatomical structures, with the segmentation results highly consistent with the ground truth. However, some of the comparative models exhibited false negatives and false positives during segmentation. For instance, both the UperNet and DeepLabV3+ models showed missed detection of key structures, with certain structures visibly absent in the visualized results. In cases where the overall cross-sectional area is smaller, the issue of structural omission becomes even more pronounced.
\section{Conclusion}
In this paper, we  propose MS-UMamba, which explores the impact of hybrid convolutional-mamba and Multi-scale feature fusion on model performance. We developed an SS-MCAT-SSM block, the ADFF Block, which is suitable for both capture local context and global context information. Additionally, we have introduced spatial attention mechanism and Monte Carlo attention mechanism into the designed MS-UMamba model. These methods further enhance the performance and feature representation of the UMamba model. Although our proposed  MS-UMamba achieves remarkable performance in the task of segmentation. However, the MS-UMamba model also had some limitations. First, with poor performance in segmenting small anatomical structures. Second, dataset samples are severely scarce. Finally, the parameters and depth of our model are very large. In the future, we will further optimize the MS-UMamba model by semi-supervised learning strategies and  lightweight method, and extend MS-UMamba model to a broader range of fetal ultrasound imaging segmentation.
\section*{Acknowledgements}
This work was supported by the National Natural Science Foundation of Guangxi under Grants 2024JJA141093 and 2023AB01361, the Basic Ability Improvement Project for Young and Middle-aged Teachers in Guangxi under Grant No. 2024KY0694, the Key Research Project of Wuzhou Medical College  under Grant No. 24WYZD03, the Wuzhou Science and Technology Plan Project under Grant No. 2023A05036,  and the National College Student Innovation and Entrepreneurship Training Program Funded Projects under Grant No. 202411354042.
\bibliographystyle{spmpsci_unsrt}
\bibliography{reference}

\end{document}